\documentclass[a4paper]{article}

\usepackage{INTERSPEECH2019}
\usepackage{comment}
\usepackage{tipa}
\usepackage{hyperref}

\title{End-to-end Text-to-speech for Low-resource Languages \\ by Cross-Lingual Transfer Learning}
\name{Yuan-Jui Chen$^{*1}$, Tao Tu$^{*1}$\thanks{*Equal contribution}, Cheng-chieh Yeh$^1$, Hung-yi Lee$^1$}

\address{
    $^1$College of Electrical Engineering and Computer Science, National Taiwan University
}

\email{\{r07922070, r07922022, r06942067, hungyilee\}@ntu.edu.tw}

\begin{document}
\maketitle

\begin{abstract}
End-to-end text-to-speech (TTS) has shown great success on large quantities of paired text plus speech data. However, laborious data collection remains difficult for at least 95\% of the languages over the world, which hinders the development of TTS in different languages. In this paper, we aim to build TTS systems for such low-resource (target) languages where only very limited paired data are available. We show such TTS can be effectively constructed by transferring knowledge from a high-resource (source) language.  
Since the model trained on source language cannot be directly applied to target language due to input space mismatch, we propose a method to learn a mapping between source and target linguistic symbols.
Benefiting from this learned mapping, pronunciation information can be preserved throughout the transferring procedure. Preliminary experiments show that we only need around 15 minutes of paired data to obtain a relatively good TTS system. Furthermore, analytic studies demonstrated that the automatically discovered mapping correlate well with the phonetic expertise.
\end{abstract}
% Index
\section{Introduction}
%TTS powerful in many aspect. But little is about low-resource

Recent research on end-to-end text-to-speech (TTS)~\cite{wang2017tacotron, van2016wavenet, shen2018natural, ping2017deep, taigman2017voiceloop, sotelo2017char2wav} has gained success in terms of human-like and high-quality generated speech.
Moreover, with regard to cloning prosody style or speaker characteristics, end-to-end TTS systems also demonstrate a powerful capability~\cite{skerry2018towards, wang2018style, jia2018transfer, lorenzo2018can, arik2018neural}.
However, training end-to-end TTS systems requires large quantities of text-audio paired data.
In order to improve data efficiency, semi-supervised training framework is proposed for Tacotron~\cite{wang2017tacotron} by leveraging non-parallel large-scale text and speech resources~\cite{chung2019semi}. 
Nevertheless, there is little discussion on end-to-end TTS for low-resource languages, where only very limited paired data are available.
 
 %old methods use high-res lang. to help
 % IPA or not
 
 Previous research on multi-lingual multi-speaker (MLMS) statistical parametric speech synthesis (SPSS) has discussed using high-resource languages to help construct TTS systems for low-resource languages. %~\cite{yu2016learning, gutkin2017uniform, li2016multi, demirsahin2018unified}. 
 Some research shows that the model trained on multiple languages can benefit from cross-lingual information and aid the adaptation to new languages using only a small amount of data~\cite{yu2016learning, gutkin2017uniform}. In their methods, linguistic inputs of each language are converted internally into language-independent representations. 
 On the contrary, in another work~\cite{li2016multi}, inputs are mapped to the International Phonetic Alphabet (IPA)~\cite{international1999handbook}, which is a unified canonical representation. The authors propose a language-agnostic model and also show that the model trained on many languages is sometimes better than the monolingual system built from small amounts of data.
 Likewise, another work indicates that training data for building a new TTS system can be reduced by pooling phonologically close languages, where a special phoneme inventory is presented for sharing as more regularities across languages as possible~\cite{demirsahin2018unified}. 
 % Prof. Lee: Although previous work demonstrates that cross-lingual transfer learning is beneficial to TTS, cross-lingual transfer learning has not been widely studied on end-to-end TTS yet. 
 Although previous work demonstrates that utilizing cross-lingual information is beneficial to TTS, this idea has not been widely studied on end-to-end TTS yet.

%our approach, high-res. help low-res. lang. with trsfr-lrng
%
%In this paper, we extend the ideas to end-to-end TTS and introduce cross-lingual transfer learning. %the ideas? which one
In this paper, we introduce cross-lingual transfer learning for low-resource languages to end-to-end TTS.
We first pretrain a TTS model by leveraging data from high-resource (source) language, and then try to adapt it to low-resource (target) languages.
To tackle input space mismatch across languages, we propose a Phonetic Transformation Network (PTN) model to discover a mapping between source and target linguistic symbols according to their pronunciation. 
The idea is similar to probabilistic phoneme mapping model~\cite{sim2008context,sim2009discriminative}, while our approach is pure deep-learning, and we use connectionist temporal classification (CTC) loss~\cite{graves2006connectionist} as the training objective.
Benefiting from the learned mapping, pronunciation information can be preserved throughout the transferring procedure. 
%Objective and subjective tests show that a few paired data on target language is required to obtain a relatively good TTS with our approach. %Lee: relatively => compare with what?　
Objective and subjective tests show that a few paired data on target language is required for our transfer learning approach to generate intelligible speech\footnote{Sound demos can be found at \url{https://henryhenrychen.github.io/CL-transfer-demo}}.
Under the scenario that input linguistic symbols of source and target languages are both phonemes, our approach is competitive with the transfer learning method which uses the handcrafted mapping based on IPA. 
Furthermore, even when lexicons of target languages are not accessible, our symbol mapping is still applicable and enables TTS to transfer from the source languages with phonemes as input to target languages with characters as input. 
Finally, analytic studies demonstrated that the automatically discovered mapping correlate well with the phonetic expertise.

\section{Proposed approach}
Given an input symbol sequence, end-to-end TTS system first transforms each symbol into a symbol embedding by an embedding matrix, and then according to the symbol embeddings, a generative model\footnote{For example, sequence-to-sequence model as in Tacotron~\cite{wang2017tacotron}.} outputs the spectrogram or raw waveform.
We can formulate end-to-end text-to-speech as
 \begin{equation}
  f_{\theta, W}:\mathcal{X}_{\mathcal{L}}\rightarrow\mathcal{Y} 
  \label{eq_tts_func}
 \end{equation}
 where $\theta$ denotes the parameters of the generative model, $W$ denotes learnable symbol embeddings, and $\mathcal{Y}$ denotes the  space of human speech. 
 $\mathcal{X}_{\mathcal{L}}$ is the text space for a specific language,  
 \begin{equation}
  \mathcal{X}_{\mathcal{L}} = \{\{s_t\}_{t=1}^{T}\ |\ \forall t\ s_t \in \mathcal{L} \ , T\in\mathbb{N}\} %因為 T  有不同長度，看來還是需要寫 T\in\mathbb{N}
% \mathcal{X}_{\mathcal{L}} = \{\{s_t\}_{t=1}^{T}\ |\ \forall t\ s_t \in \mathcal{L} \  \},
  \label{txt_space}
 \end{equation}
 where $\mathcal{L}$ is the linguistic symbol set for this language, and $T$ is the length of the input symbol sequence. 
 Our goal is to construct TTS systems for low-resource (target) languages by transferring knowledge from high-resource (source) language. 
 We can directly use $\theta_{src}$ learned from source language to initialize the training of $\theta_{tgt}$ on target language because both $\theta_{src}$ and $\theta_{tgt}$ take embeddings as input and generate speech\footnote{Here $src$ and $tgt$ stand for source and target, respectively.}.
 However, the same idea cannot be directly applied to $W_{src}$ and $W_{tgt}$. 
 An obvious problem is that ${s}_{src}$ and ${s}_{tgt}$ come from different symbol sets, i.e., $\mathcal{L}_{src}\neq\mathcal{L}_{tgt}$.
 To deal with the input space mismatch problem during the transferring procedure, we present two naive baselines and propose a novel transfer learning approach which utilizes a learned mapping between ${s}_{src}$ and ${s}_{tgt}$.

 \begin{figure}[t]
 \centering
 \includegraphics[width=8cm]{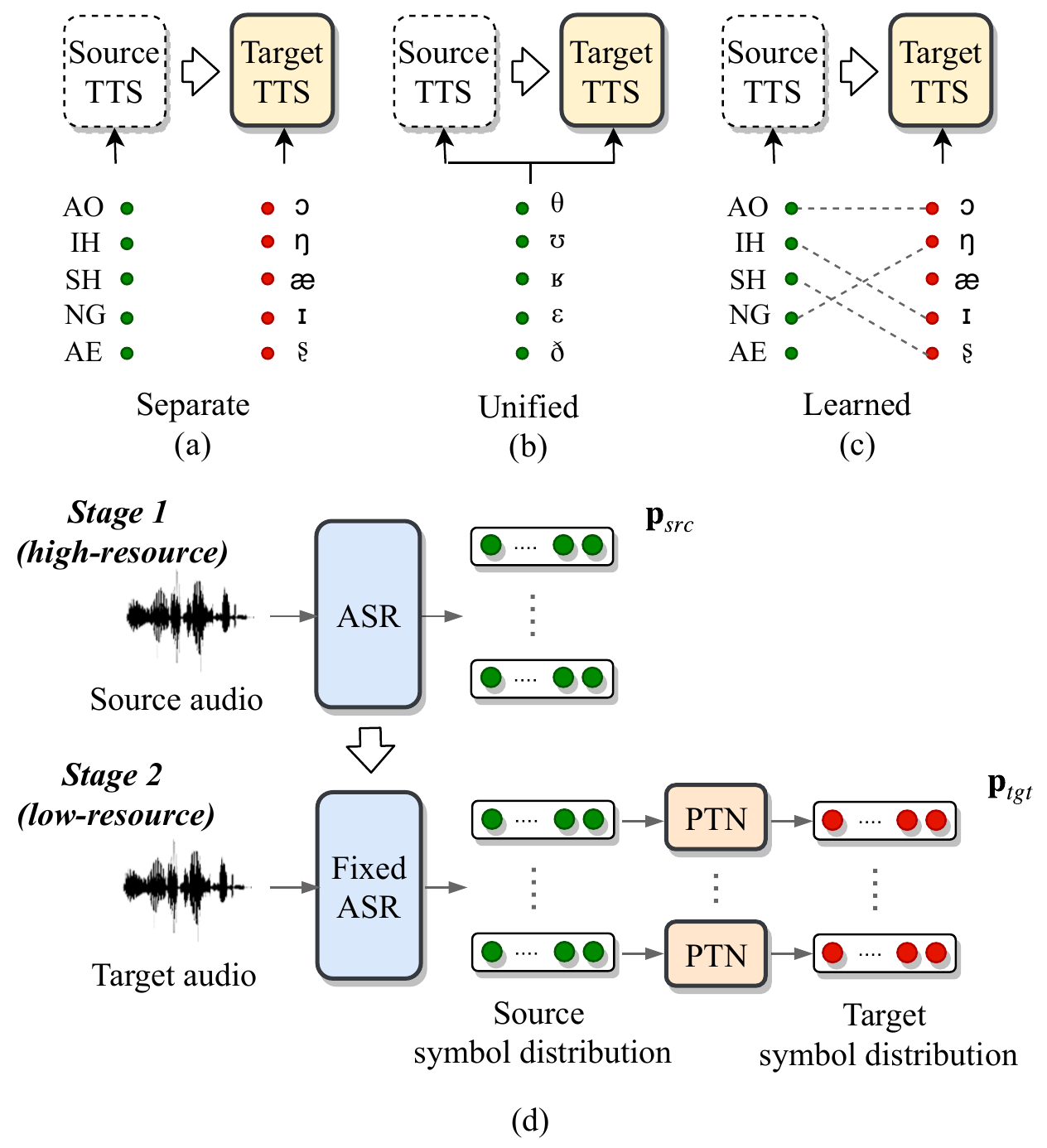}
 \caption{Approaches to transfer TTS model from source language  to target language. 
 (a) separate symbol space,
 (b) unified symbol space,
 and (c) learned symbol space. 
 (d) the training scheme of phonetic transformation network (PTN) for obtaining the learned symbol space.}
 \label{fig:tts_asr}
\end{figure}
 \subsection{Separate symbol space}　
\label{subsec:separate}
%Due to input space mismatch, source symbol embeddings $W_{src}$ cannot be transferred with the sequence-to-sequence model to target language. %這句前面講過了
%A naive approach is to regard linguistic symbol sets for source and target language separate.
%Therefore, the TTS system for target language can be derived by jointly finetuning the pretrained sequence-to-sequence model and new target symbol embeddings $W_{tgt}$.
The first approach simply considers linguistic symbol sets for source and target language as two different symbol sets.
In this approach, $\theta_{tgt}$ is derived by finetuning $\theta_{src}$, but the target symbol embeddings $W_{tgt}$ is learned from scratch. 
%While this method seems naive, it is indeed a strong baseline.

 \subsection{Unified symbol space}
\label{subsec:unified}
However, some of the sound units are shared by different languages. If we discard $W_{src}$ and train new $W_{tgt}$, some useful pronunciation information learned previously may be lost.
This can be resolved by mapping $\mathcal{L}_{src}$ and $\mathcal{L}_{tgt}$ to a unified symbol set $\mathcal{L}_{uni}$, where the mapping is handcrafted and relies on linguistic expertise.
%In this way, we can perform both pretraining and finetuning with the same unified symbol embeddings $W_{uni}$. %W_{uni}後面都沒有出現過了，不用再多用一個符號
In this way, we can use $W_{src}$ to initialize $W_{tgt}$ because they have the same set of input symbols $\mathcal{L}_{uni}$. 
Note that this method necessitates experts to design symbol mapping for the source and the target language.
This kind of mapping is not always available especially when the symbol set of one language is phoneme, while the other is character.

 \subsection{Learned symbol space}
\label{subsec:map}

 % Introduce why we learn mapping (PTN)
 %Unified symbol representation relies on expert knowledge to design symbol mapping, which is against our low-resource setting.
 %Thus, we want to discover automatically the mapping between a source symbol and another target symbol that represent the same sound.
 To preserve pronunciation information during transferring while not using linguistic expertise, we propose Phonetic Transformation Network (PTN), a model that can automatically learn how to map source symbols to target symbols according to their sounds.
 
\subsubsection{Phonetic transformation network}
% Explain why we need ASR first. (because we'd like to find patterns in voice across language) 
% Detail the training process
%    - stage 1 (ASR for source language)
%    - stage 2 (PTN training)
First, we pretrain an automatic speech recognition (ASR) system on source language, as illustrated in stage 1 of Figure~\ref{fig:tts_asr}(d).
The ASR system learns to output symbol (phoneme) sequence of the source language by CTC loss. 
%The ASR system learns to transcribe the source language to symbol (phoneme) sequence by CTC loss.
Afterward, we fix the pretrained ASR system and concatenate our proposed PTN model with it.
PTN can be formulated as 
\begin{equation}
  h: \mathbf{p}_{src}\mapsto\mathbf{p}_{tgt}
  \label{eq_ptn_func}
\end{equation}
where $\mathbf{p}_{src}$ and $\mathbf{p}_{tgt}$ are probability distributions over $\mathcal{L}_{src}$ and $\mathcal{L}_{tgt}$ for a specific timestep. 
In our case, $\mathbf{p}_{src}$ is also the ASR output symbol posteriorgram.
%We can intuitively regard the ASR output posteriorgram as $\mathbf{p}_{src}$ where each dimension represents the similarity ratio between the sound of the corresponding symbol in the source language and the input audio frame. %???
The concatenation of the pretrained ASR system and PTN is then further trained on the target language data by maximizing the log-likelihood of target symbol labellings (phonemes or characters of the target language) using CTC loss, as illustrated in stage 2 of Figure~\ref{fig:tts_asr}(d). %, where the goal of PTN is to learn the mapping between $\mathcal{L}_{src}$ and $\mathcal{L}_{tgt}$.
In stage 2, the parameters of the ASR system are fixed, so what PTN has learned is to find the most possible target symbols given the ASR output which are source symbols. 
%Afterwards, we combine PTN model, whose goal is to learn the mapping between $\mathcal{L}_{src}$ and $\mathcal{L}_{tgt}$, with the pretrained ASR to recognize target audio.
Since the pretrained ASR system is capable of transcribing an audio frame in target language into a posteriorgram of source symbols, the training in stage 2 enables PTN to learn a strategy to convert the symbols (phoneme) of the source language into the symbols (phonemes or characters) of the target language.
%transcribed posteriorgram to the probability distribution in the target language.
%Since each dimension of ASR output posteriorgram represents the degree of sound from each source symbol, the strategy that PTN learns is to find the corresponding target symbol given different ratio of sounds from source symbols.

\subsubsection{Symbol mapping discovery}
\label{subsec:map-discovery}
With PTN,  we can derive the most similar target symbol to a certain source symbol according to their sound. 
Given the i-th source symbol $s_{src}^{i}$, we can simply pass a one-hot vector $\mathbf{o}_{i}$, whose i-th dimension is marked as 1, to PTN. If the sound of $s_{src}^{i}$ is shared among source and target language, PTN will convert $\mathbf{o}_{i}$ to a target symbol with high probability.
Accordingly, we can map each source symbol to a target symbol by the following formulation.
 \begin{equation}
  map(s_{src}^{i})=
  \begin{cases}
  s_{tgt}^{j}\ & if \ \ h^{j}(\textbf{o}_{i}) > \xi,\ j=\underset{k}{\arg\max} \ h^{k}(\textbf{o}_{i}) \\
  \text{None}\ & otherwise
  \end{cases}
  \label{eq_mapping_func}
 \end{equation}
where $h^{k}(\cdot)$ denotes the k-th output dimension of PTN $h(\cdot)$, $s_{tgt}^{j}$ denotes j-th symbol in the target language and $\xi$ is the transformation threshold.
Once obtaining the mapping, we can transfer the embedding weight of a source symbol to its corresponding target symbol. 
If a target symbol is mapped by many source symbols, we transfer the embedding weight of the one with the highest probability.
For those symbols in the target language which are not mapped by any source symbol, their embedding weights are still learned from scratch. %不知道需不需要這句 (空間不夠可拿掉)

\subsubsection{Symbol mapping discovery}
\label{subsec:map-discovery}

 Because PTN is capable of converting a posteriorgram of source language into probability distribution in target language, it can be used to derive the cross-lingual symbol mapping function.
 %For each symbol $s_{src}$ in source language, we obtain its corresponding mapped target symbol $s_{tgt}$ relying on the mapping function $q(\cdot)$.
 For i-th symbol $s_{src}^{i}$ in source language, we obtain its corresponding mapped target symbol relying on $h_{\psi}(\cdot)$.
 Concretely speaking, we pass a one-hot vector $\mathbf{o}_{i}$, whose i-th element is marked as 1, to PTN. Then, the target symbol with the highest value in the output probability $h_{\psi}(\textbf{o}_{i})$ is regarded as the mapped target one. 
 \begin{equation}
 \begin{aligned}
  &map(s_{src}^{i})=s_{tgt}^{j} \\  
  j&=\underset{k}{\arg\max} \ h_{\psi}^{k}(\textbf{o}_{i})
  \label{eq_mapping_func}
 \end{aligned}
 \end{equation}
 where $s_{src}^{i}$ and $s_{tgt}^{j}$ denote the i-th source symbol and the j-th target symbol, $h_{\psi}^{k}(\cdot)$ is the k-th element of $h_{\psi}(\cdot)$ and $\xi$ is the transformation threshold. 
 Accordingly, we can transfer the source symbol embeddings $W_{src}$ to target language $W_{tgt}$.
 For the target language symbol not mapping to any source language symbol, their corresponding embedding weights are still learned from scratch.
 
\begin{comment}
 by initializing the target symbol embeddings with the corresponding source one.
 If there are more than one source symbols are mapped to a target symbol, we keep the one having the highest probability. Also, we randomly initialize embeddings for target symbols not mapped by any source symbol.
  \begin{equation}
  map(s_{src}^{i})=
  \begin{cases}
  s_{tgt}^{j}\ & if \ \ h_{\psi}^{j}(\textbf{o}_{i}) > \xi,\ j=\underset{k}{\arg\max} \ h_{\psi}^{k}(\textbf{o}_{i}) \\
  \text{None}\ & otherwise
  \end{cases}
  \label{eq_mapping_func}
 \end{equation}
\end{comment}

\section{Implementation}
\subsection{TTS model}
In this work, we adopt original Tacotron architecture~\cite{wang2017tacotron} as our end-to-end TTS model, which has an encoder-decoder architecture with attention mechanism. 
Spectral analysis setting is also the same as theirs~\cite{wang2017tacotron} in the paper.
Since our goal is to study transfer learning in the small-data regime, we simply use Griffin-Lim~\cite{griffin1984signal} as the waveform synthesizer and leave exploring other architectures~\cite{shen2018natural, wang2018style} as our future work. 
%In addition, we choose Griffin-Lim algorithm~\cite{griffin1984signal} to convert the predicted spectrograms to waveforms for faster experiment cycles. 

\subsection{ASR and PTN model}
%short version
A pure-CNN model is adopted for our ASR system, which is modified from the previous work~\cite{krishna2018study}. A pyramidal recurrent neural network (RNN) model \cite{chan2016listen} was also experimented, whereas we find it performed not as expected in preliminary studies. 
We conjecture that RNN with multiple layers has learned strong language model on source language, which laid constraints on model's outputs and hindered the training of subsequent PTN. 

As for PTN, it is composed of 3-layer fully connected layers with ReLU activation function. Dropout is also applied with 0.4 dropout rate for each layer. 
\section{Experiments}
To verify whether TTS model can benefit from cross-lingual transfer learning and generate clear speech with small amounts of data, both objective and subjective tests are conducted.
For the objective tests, we use google's cloud speech-to-text API to recognize the generated speech and use the character error rate (CER) as the measurement metric for clarity. Additionally, we also use mel-cepstral distortion (MCD)~\cite{kominek2008synthesizer} for evaluation, which measures the distance between synthesis and ground truth in the space of mel-frequency cepstrum --- the smaller the better.
For subjective measurements, mean opinion score (MOS) tests are run for naturalness assessment.
%and the model trained from scratch where all network weights are randomly initialized

For simplicity, "phn2phn" denotes the situation using phoneme as input in both source and target languages, and "phn2char" denotes the situation using phoneme input in source language but character input in target languages.
%As for the model trained from scratch, we denote it as "Scratch"
Likewise, we denote the model that transfers with separate symbol space, unified symbol space and learned symbol space by "Separate", "Unified" and "Learned", respectively. 

\subsection{Data setup}
\subsubsection{Source language}
In our experiments, English was selected as our high-resource language. For pretraining an initial TTS model, LJ Speech Dataset~\cite{ito2017lj} is used, which is a public domain speech dataset consisting of around 24 hours of text speech paired data. 
As for ASR training in Section~\ref{subsec:map}, we use the LibriSpeech Dataset~\cite{panayotov2015librispeech}, which is an ASR corpus based on public domain audio books. 
The training set of 100 hours clean speech and the clean development set are utilized for training and early stopping.

\subsubsection{Target language}
Mandarin, German, and French are chosen as the target languages.
%, which contains roughly 37 minutes of read speech with transcription by a single female speaker. 
An internal corpus recorded by a female speaker is used for Mandarin experiments.
The German data derives from the German LibriVox corpus which is organized by M-AILABS~\cite{de_de}. Data from a female speaker Eva K is used.
As for French, we use the data from a female speaker FR010 in the GlobalPhone collection~\cite{schultz2002globalphone}, which only consists of approximate 18 minutes paired data. 
We split the data into training and testing sets as illustrated in Table~\ref{tab:data-table}\footnote{Mandarin and German use the same test sets for both CER and MCD measurements. However, since there is very few French data and MCD test needs ground-truth audio, we randomly select needed training data and leave the rest for testing. This procedure is run three times and the average score is reported.}.
\begin{table}[h]
  \caption{Data statistic of target languages}
  \label{tab:data-table}
  \centering
  \begin{tabular}{ l@{}l  c c}
    \toprule
    \multicolumn{2}{c}{\textbf{Language}} & 
    \multicolumn{1}{c}{$\textbf{Train}_\text{(minutes)}$} &
    \multicolumn{1}{c}{$\textbf{Test}_\text{(utterances)}$} \\
    \midrule
    $\text{Mandarin}$ & & $30$ & $250$ \\
    $\text{German}$ & & $30$ & $120$  \\
    $\text{French}$ & & $15$ & $100$ \\
    \bottomrule
  \end{tabular}
\end{table}

\subsection{Experimental setup}
 The initial TTS model is obtained by pretraining on source language for 10k parameter updates. 
 %Lee: 我沒看過"10000 mini-batch"這樣的講法，應該是 with 10,000 parameter updates
% For all transfer methods, we utilize the same initial TTS model.
 For all transfer learning methods, we continue training on the target language pair with the same initial TTS model parameters. 
 
 %In "Separate" (Section~\ref{subsec:separate}), source and target languages have their own symbol embedding matrix. Thus, embedding matrices for target languages are randomly initialized according to the normal distribution with 0 mean and 0.3 standard deviation. 
 In "Separate" (Section~\ref{subsec:separate}), embedding matrices for target languages are randomly initialized according to the normal distribution with 0 mean and 0.3 standard deviation. 
 In "Unified" (Section~\ref{subsec:unified}), all symbols are mapped to IPA. 
 Accordingly, for each symbol of the target language, we initialize its embedding weight from the source symbol that shares the same IPA representation.
 The embeddings for the remaining symbols are randomly initialized as explained for "Separate".
 As for "Learned" (Section~\ref{subsec:map}), ASR model in stage 1 is pretrained on source language for 300k parameter updates and the best model is selected by the development set. 
 %Lee: 同上，"300000 min-batch"應改掉
 The training data for PTN is the same for finetuning the TTS model on the target languages.
 The transformation threshold $\mathbf{\xi}$ is set to 0.4 for all target languages. 
 Finally, embeddings for target symbols are initialized in the same way as "Unified", except that the mapping is now learned automatically.
 %Lee: 這裡的三個方法，跟後面 fig 2 和 3 上面的方法名稱是不一致的，應該統一

\subsection{Experiment results}
\subsubsection{Objective tests}
\begin{figure}[h]
  \centering
  \includegraphics[width=.48\textwidth]{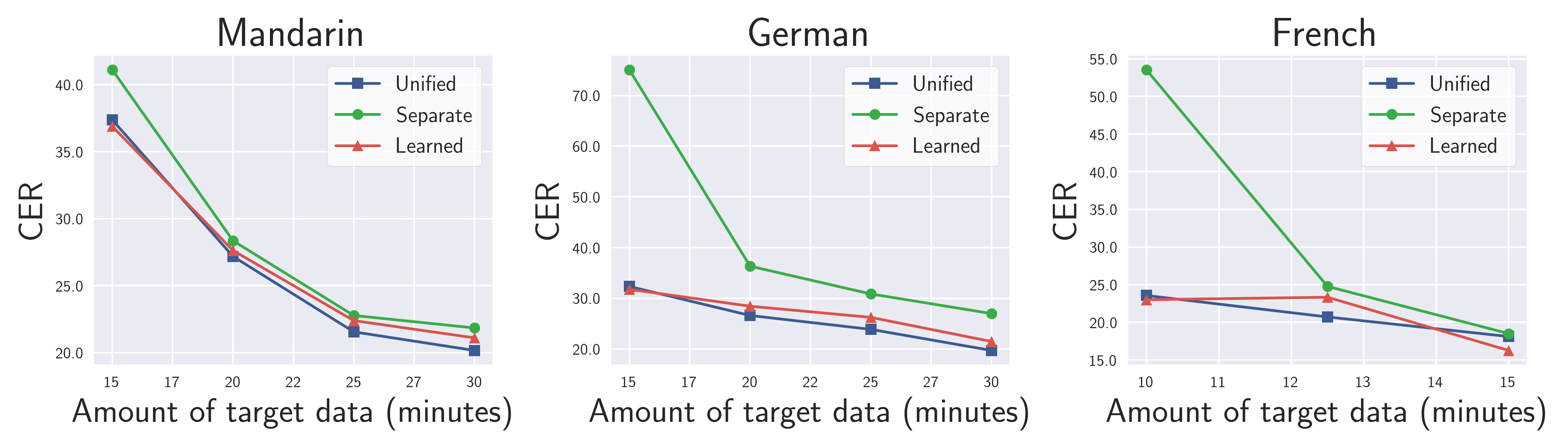}
  \caption{Results of CER under "phn2phn" scenario.}
  \label{fig:cer_phn2phn}
\end{figure}

\begin{figure}[h]
  \centering
  \includegraphics[width=.48\textwidth]{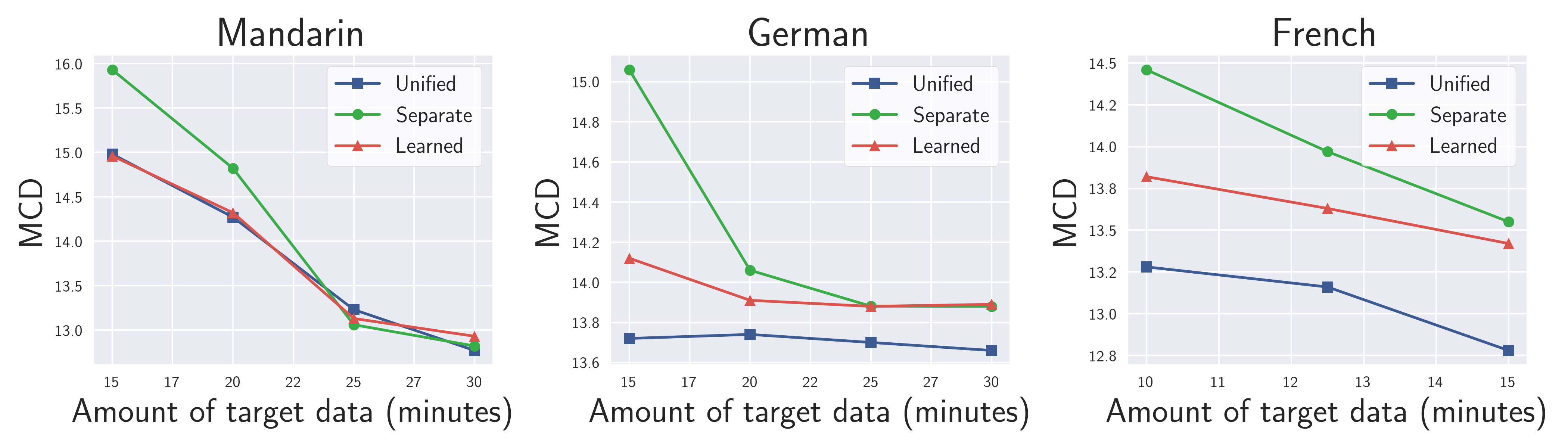}
  \caption{Results of MCD tests under "phn2phn" scenario.}
  \label{fig:mcd_phn2phn}
\end{figure}

First of all, we show the results in the situation "phn2phn", where lexicons for target languages are accessible.
The CER results are shown in Figure~\ref{fig:cer_phn2phn}. We can see that for any language and any amount of used target data, "Unified" and "Learned" consistently outperform "Separate", which implies that transferring knowledge with the symbol (phoneme) mapping is beneficial.
When the size of target data decreases, "Separate" deteriorates the most and "Learned" sticks with "Unified". 
This also indicates that the mapping information is especially effective under very scarce data circumstances and that our learned mapping is competitive with the one based on IPA. 
Besides, we can notice from Figure~\ref{fig:mcd_phn2phn} that the results of MCD tests also align with the results of CER.
The model trained from scratch, where all network weights are randomly initialized, is experimented. However, even if all training data is used, it cannot produce understandable speech and results in CER larger than 80\% for every language. Thus, we do not plot its results.

\begin{figure}[h]
  \centering
  \includegraphics[width=.48\textwidth]{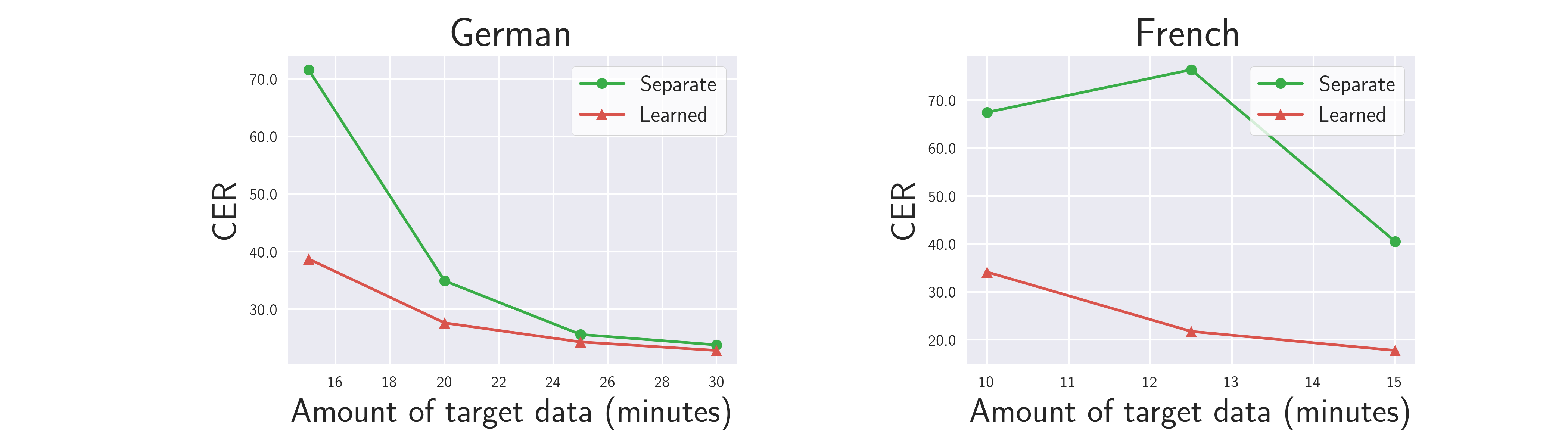}
  \caption{Results of CER under "phn2char" scenario.}
  \label{fig:cer_phn2char}
\end{figure}

In addition, "phn2char" setting is also investigated.
Because under such setting, the input symbols of target languages are characters, "Unified" approach is not applicable.
In Figure~\ref{fig:cer_phn2char}, a large gap between "Learned" and "Separate" can be observed on German and French\footnote{Because the characters of Mandarin correspond to syllable instead of phoneme, "phn2char" is not reasonable for Mandarin, so its performance is not presented here.}.
This shows that our proposed method performs well even when source symbols are phoneme-level and target symbols are character-level.

%Interestingly, comparing "Learned" in Figure~\ref{fig:cer_phn2phn} and Figure~\ref{fig:cer_phn2char} , we find that in French, the model under "phn2phn" performs better than under "phn2char", while in German, an opposite result is observed. 
%Because phoneme transcriptions for German data are not perfectly cleaned due to our unfamiliarity to German, we hypothesize the errors in transcriptions lead to such result. 
%Furthermore, it also indicates that when lexicons for a language is not accessible or not clean, our proposed transfer learning method is effective and comes in handy. 
\vspace{-1.8\baselineskip}
\vspace{-1.6\baselineskip}

\subsubsection{Subjective tests}
\begin{table}[h]
  \caption{Mean Opinion Score (MOS) ratings with 95\% confidence intervals for \textbf{naturalness}.}
  \label{tab:MOS}
  \centering
  \begin{tabular}{ l@{}l c c }
    \toprule
    \multicolumn{2}{c}{} & 
    \multicolumn{2}{c}{\textbf{MOS score}} 
    %\multicolumn{1}{c}{}
    \\
    \multicolumn{2}{c}{\textbf{Method}} &
    \multicolumn{1}{c}{$25$ minutes} &
    \multicolumn{1}{c}{$15$ minutes}
    \\
    \midrule
    Ground Truth & & %Lee: Natural 這個用法比較少見
    \multicolumn{2}{c}{$4.89\pm0.045$} \\
    %Ground-Truth & & $4.91\pm0.35$ & $4.86\pm0.52$ \\ 
    Separate & & $3.94\pm0.085$ & $2.90\pm0.176$    \\
    Unified & & $4.01\pm0.085$ & $3.48\pm0.119$     \\
    Learned & & $3.99\pm0.086$ & $3.46\pm0.117$     \\
    Scratch & & $1.39\pm0.153$ & $1.26\pm0.094$     \\
   %Lee: "Learned" 不需要粗體
    \bottomrule
  \end{tabular}
\end{table}

%\begin{table}[h]
%  \caption{5-scale mean opinion score evaluation for %\textbf{completeness}}
%  
%  \label{tab:MOS}
%  \centering
%  \begin{tabular}{ c@{}l c c }
%    \toprule
%    \multicolumn{2}{c}{} & 
%    \multicolumn{2}{c}{\textbf{MOS score}} 
%    %\multicolumn{1}{c}{}
%    \\
%    \multicolumn{2}{c}{\textbf{Method}} &
%    \multicolumn{1}{c}{$25$ minutes} &
%    \multicolumn{1}{c}{$15$ minutes}
%    \\
%    \midrule
%    Truth & & $ 4.93\pm0.33 $ & $ 4.90\pm0.49 $ \\ 
%    Unified & & $ 4.77\pm0.54 $ & $ 4.40\pm0.89 $     \\
%    Separate & & $ 4.71\pm0.58 $ & $ 2.70\pm1.56 $    \\
%    Learned & & $ 4.75\pm0.55 $ & $ 4.54\pm0.74 $     \\
%    Scratch & & $ 1.41\pm0.84 $ & $ 1.11\pm0.33 $     \\
%    
%    \bottomrule
%  \end{tabular}
%\end{table}
To further examine the impact of target data size on the quality of generated speech, we conduct a series of MOS tests.
We use 25 minutes and 15 minutes Mandarin paired data for this test under "phn2phn" setting.
The model trained from scratch (denoted by "Scratch") is also measured for comparison. 
In MOS tests, 40 subjects were asked to rate the naturalness for the given speech audio and 80 audio of unseen utterances were used for testing. Each utterance received 5 ratings at least.
After listening to each stimulus with headphone, the subjects were asked to rate the naturalness in a five-point Likert scale score (1: Bad, 2: Poor, 3: Fair, 4: Good, 5: Excellent).

From Table~\ref{tab:MOS}, we can observe that when 25 minutes of paired data is used, three transfer learning methods "Separate", "Unified" and "Learned" perform almost the same and all of them outperform "Scratch". 
When training data is reduced to 15 minutes, "Separate" degrades obviously, which is consistent with the discovery of previous objective tests.
The results show that given a few but still sufficient paired data (25 min) on the target language, three transfer learning approaches can benefit from pretraining and generate intelligible speech. 
When paired data becomes fewer (15 min), our proposed approach "Learned" is comparable to "Unified" and gives promising results without using background linguistic expertise.

\subsection{Symbol mapping studies}
\begin{table}[h]
  \caption{Precision and recall of found mapping on 15-minute target data.}
  \label{tab:map}
  \centering
  \begin{tabular}{ l@{}l  c c c}
    \toprule
    \multicolumn{2}{c}{\textbf{Mapping}} & 
    \multicolumn{1}{c}{\textbf{Precision}} &
    \multicolumn{1}{c}{\textbf{Recall}} &
    \multicolumn{1}{c}{$\textbf{Recall}_\text{random}$} \\
    \midrule
    $\text{EN} \rightarrow \text{DE}$ & & $82.6\%$ & $63.3\%$ & $3.4\%$ \\
    $\text{EN} \rightarrow \text{FR}$ & & $73.7\%$ & $56.0\%$ & $4.0\%$\\
    $\text{EN} \rightarrow \text{ZH}$ & & $64.7\%$ & $47.8\%$ & $4.5\%$\\
    %Phonetics & & $33.6\%$ & $53.4\%$ \\
    \bottomrule
  \end{tabular}
\end{table}
%In this part, we evaluate the effectiveness of our learned phoneme mapping by precision and recall.
In this part, we show that our learned symbol mappings are reasonable and evaluate them according to IPA under "phn2phn" setting. 
%In this part, we show that the learned phoneme mappings are reasonable.
If one source phoneme and its learned corresponding target phoneme share the same IPA representation, we regard this learned mapping correct.
To calculate recall score, we derive total correct mappings from the overlap of two language phoneme sets after being mapped to IPA.  
%, which is defined as $\{\text{IPA}(s_{src})\ |\ \forall s_{src} \in \mathcal{L}_{src}\} \cap \{\text{IPA}(s_{tgt})\ |\ \forall s_{tgt} \in \mathcal{L}_{tgt}\}$.
%Lee: 這句看不懂
For the sake of comparison, we also show the recall score in the case that each source phoneme is randomly mapped to a target phoneme in the overlap.
From Table~\ref{tab:map}, we can observe that our method retrieves highly informative mapping and is far better than random guessing. Besides, we can notice that our method performs better on German and French than on Mandarin, which may result from the similarity to the source language, English. 
Despite relatively low recall score on Mandarin, our method still discovers some mappings between two similar-sounding phonemes which have different IPA representations. 
For example, symbol~$<$\textipa{\:s}$>$ and symbol~$<$\textipa{S}$>$ are mapped. Although they are not identical according to IPA, their pronunciations are quite alike and similar to "sh" in English.
For more details about the learned mapping please refer to the demo page.

\section{Conclusion}
 In this paper, we explored cross-lingual transfer learning in end-to-end TTS for low-resource languages. We proposed an approach to discover cross-lingual symbol mapping for helping model better transferred with knowledge learned previously from abundant source data. Experiment results show that our method enables the model to produce far more natural-sounding speech than the model trained only on target data and achieves promising results compared with the method using strong linguistic background expertise.

\bibliographystyle{IEEEtran}
\bibliography{ref}

\end{document}